\documentclass[10pt,twocolumn,letterpaper]{article}

\usepackage{btas}
\usepackage{times}
\usepackage{epsfig}
\usepackage{graphicx}
\usepackage{amsmath}
\usepackage{amssymb}
\usepackage{balance}

\usepackage{subfigure}
\usepackage[utf8]{inputenc}
\usepackage[T1]{fontenc}

   \DeclareGraphicsExtensions{.pdf,.jpeg,.png}
\usepackage{url}
\usepackage{booktabs}

\usepackage{bm}
\usepackage{amsfonts}
\usepackage{microtype}

\btasfinalcopy 

\ifbtasfinal\fi

\begin{document}
%
\title{Hybrid Template Update System for Unimodal Biometric
Systems}

\author{Romain Giot and Christophe Rosenberger\\
{Universit\'e de Caen, UMR 6072 GREYC}\\
{ENSICAEN, UMR 6072 GREYC}\\
{CNRS, UMR 6072 GREYC}\\
{\tt \small\{romain.giot,christophe.rosenberger\}@ensicaen.fr}\\
\and
Bernadette Dorizzi\\
{Institut Mines-Telecom sudParis}\\
{UMR 5157 SAMOVAR}\\
{\tt\small bernadette.dorizzi@it-sudparis.eu}
}


%


\maketitle
\thispagestyle{empty}

\begin{abstract}
Semi-supervised template update systems allow to automatically
take into account the intra-class variability of the biometric
data over time.
Such systems can be inefficient by including too many
impostor's samples or skipping too many genuine's
samples.
In the first case, the biometric reference drifts
from the real biometric data and attracts more often impostors.
In the second case, the biometric reference does not evolve 
quickly enough and also progressively drifts from the real biometric data.
We propose a hybrid system using several biometric sub-references in
order to increase performance of self-update systems by reducing
the previously cited errors.
The proposition is validated for a keystroke-dynamics
authentication system  (this modality
suffers of high variability over time)
on two consequent datasets from the state of
the art.
\end{abstract}

\section{Introduction}

Biometric authentication systems allow authenticating individuals
by comparing a query provided by the claimant to its biometric
reference. 
Depending on the result of this comparison, the claimant is
accepted (the system asserts he/she owns the identity he/she claims) or
rejected (the system does not assert he/she owns the identity
he/she
claims).
Usually, the biometric reference is created during the enrollment
phase by providing one or several captures.
However, most biometric modalities are not permanent and system
performance decreases with time.
To overcome this drawback, it is possible to re-enroll the user at a fixed
time.
Sadly, this method has a high cost because it needs time and operators.
Enrollment period may also be on a too short timespan to collect
enough intraclass variabilities to represent the user as best as
possible.

The aim of semi-supervised template update systems is to address
these issues by automatically updating the biometric reference of
individuals while they use the system.
The update system only uses information from the query and from the
biometric recognition system.
Semi-supervised template update is an active field of research mainly studied for morphological
modalities, whereas they are less subject to variabilities than
the behavioral ones.
As such systems can include impostor's samples in the updated
biometric reference,
the biometric reference can progressively deviates from
the owner's real biometric data and the system attracts more impostors.
There are two kinds of systems in the literature, the self-update
systems~\cite{rattani2011temporal,seeger2011how,giot2012performance}
and the co-update systems~\cite{roli2007template,bhatt2011co}.
Self-update systems allow unimodal system to update the reference
automatically after collecting unlabelled data.
Their main drawback is the fact that they are not able to attract
genuine samples to much dissimilar than the original reference
one~\cite{rattani2008biometric},
so they miss
several samples during the update procedure.
Co-update systems allow using multimodal systems in order to
attract these forgotten samples because one biometric reference is
updated based on the classification result of the complementary classifier
related to the other biometric reference linked to the other
modality.

We propose a new hybrid template update system.
There are several components in a template update system.
 Our hybrid system is not clearly based on the optimisation of one 
 particular component.
  We can see it as both a modification of the way of
  representing the biometric reference, and the way of updating
  the user's gallery.
A user is represented by several biometric
sub-references evolving in parallel by using different template
update methods.

The contributions of this work are the following ones:
(1) we propose an original hybrid template update
	system scheme performing better than the classic
	self-update system from the state of the art.
	This is an hybrid system because (a) it operates fusion as in
	co-update systems, whereas it is a self-update system,
	and (b) user's biometric reference is composed of several
	biometric sub-references;
%
(2) we propose two metrics in order to evaluate the
	efficiency of template update systems over 
	several sessions;
(3) we evaluate the method with a dataset providing more
	samples per user than most studies of the state of the
	art.

The paper is organized as follows.
Section~\ref{sec_previous_work} gives a quick overview of the
recent works on template update.
Section~\ref{sec_proposal} presents the template update
architecture we propose, as well as two new evaluation metrics.
Section~\ref{sec_protocol} presents the selected protocol to
evaluate our contribution.
Section~\ref{sec_results} presents the experimental results and
Section~\ref{sec_conclusion} concludes this communication.

\section{Similar and Recent Works}
\label{sec_previous_work}
In this section, we present the most recent works in template
update.
Bhatt~\emph{et al.} present a co-update method allowing to update
two related classifiers in an online way~\cite{bhatt2011co}.
The SVM boundary decision is updated in a semi-supervised way.
Namely, if a classifier returns with a high probability that a sample
corresponds to a particular label, whereas the second classifier
disagrees, this
new sample is used for an online update of the second
classifier.
The authors show, on a face recognition problem, that their system
improves performance both in accuracy and in computational time.
The validation is done on an aggregated database of 1833 subjects
providing 20150 images.
There is an average of 11 images per individual, which can be
considered as small for a template update study.

Rattani~\emph{et al.} present self-update and co-update for
biometric modalities where a biometric sample can be used as a
biometric reference~\cite{rattani2011temporal}.
Such kind of information can be irrelevant for some biometric
modalities, like keystroke dynamics, which are not consistent
enough to work
with only one sample.
They analyse the behavior of the updating method by representing
the samples as nodes in a graph and similarities as edges between
nodes.
They show that the graph can contain independent sub-graphs.
Samples from a sub-graph cannot attract samples from other
sub-graphs as they are too much dissimilar.
The samples present in other sub-graphs contain more variabilities
but will not be used in the template update system.
Co-update allows attracting these samples.
The study is done with 40 users providing each 50 samples (of face
and fingerprint) on 5
sessions captured on 1.5 years.

Seeger and Bours list various factors used to specify an
evaluation scenario of a template update
system for keystroke dynamics~\cite{seeger2011how}.
Note that most of the results of this paper are also relevant for
other modalities, and therefore this paper is worth reading.
The authors show that different evaluation scenarios give different
interpretation of the template update system performance.
This is a problem, because almost no template update study uses the same
kind of scenario and because most studies do not explain which
scenario configuration has been chosen.

Giot \emph{et al.} raise some questions, without answering them,
about the evaluation of template update
systems~\cite{giot2012performance}.
They show that, in addition to the scenario parameters presented
in~\cite{seeger2011how}, most studies also present a great variability in the
way of computing the performance of the template update system.
They use three different ways encountered in the template update literature to
evaluate the performance of a keystroke dynamics template update
system using exactly the same set of scores.
They show that different
interpretations can be proposed whereas the scores are identical.
These recent works assert the fact it is necessary to clearly specify the
way of computing the performances, and the need of
standardised evaluation procedures.

\section{Proposed Semi-supervised Template Update Method and Evaluation Metrics}
\label{sec_proposal}
  This section presents the proposed template update component and
  associated evaluation metrics.

\subsection{Template update based on multiple galleries evolution }
Here are some definitions for the paper.
A \emph{user's gallery} is a set of biometric samples used to represent a
user, while
a \emph{biometric reference} is a model representing a user and has
been computed with the samples of its gallery.
These two different terms are both named model, biometric reference,
or template in the literature.

\begin{figure*}[!tb]
    \centering
    \includegraphics[width=.6\linewidth]{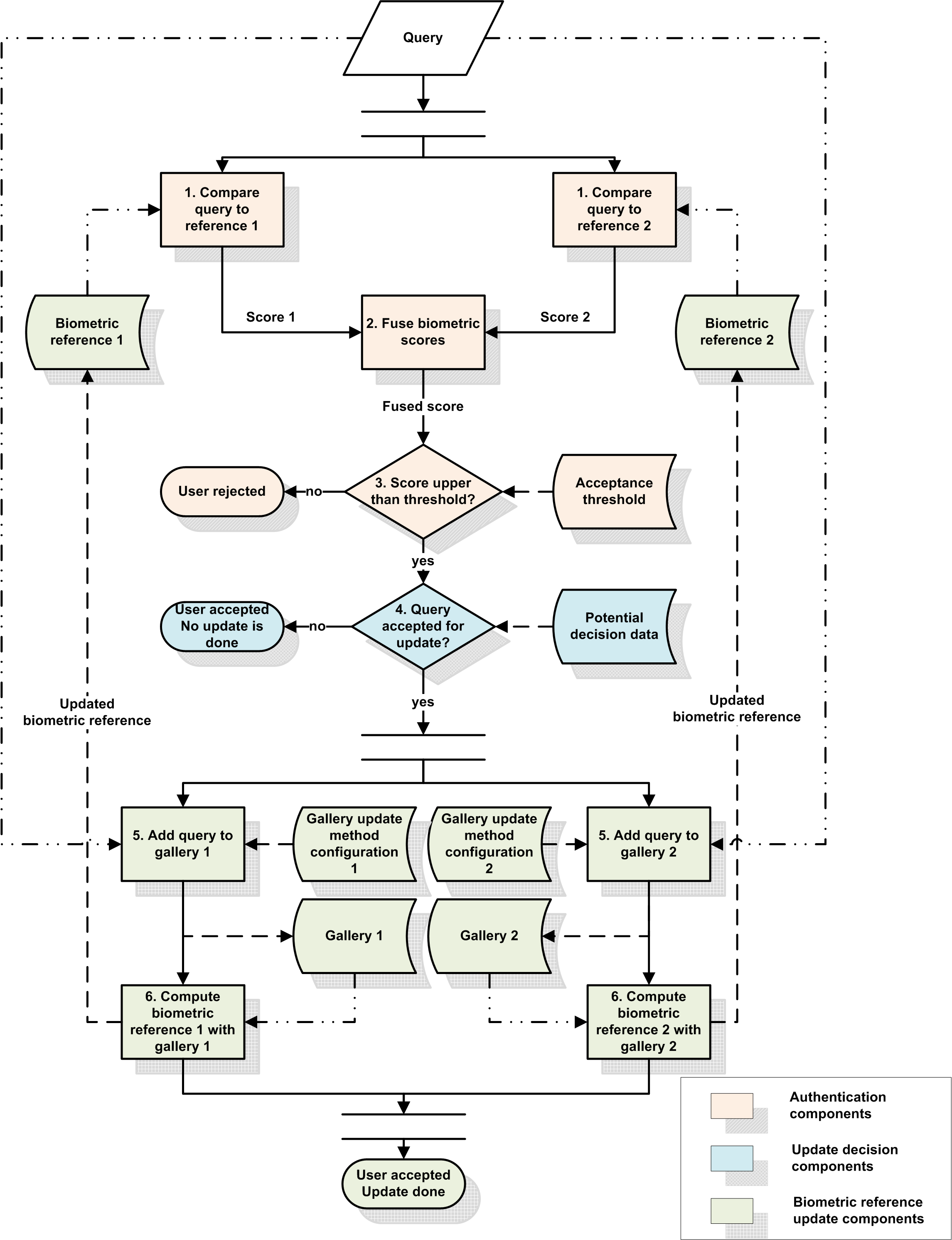}
    \caption{Workflow of the hybrid template update system (when
    two template update systems are used, in an online scenario).}
    \label{fig_schema}
  \end{figure*}

  Our contribution is inspired by the co-update
  systems~\cite{roli2007template,bhatt2011co},
  although we use a mono-modal system, and
  the various works on gallery
  update~\cite{scheidat2007automatic,kang2007crk}.
  In all previous works, the biometric reference of the user is
  unique because a user is represented by only one gallery or one sample
  or one model.
  But in our work, the biometric reference is composite: this
  \emph{biometric meta-reference} contains
  several \emph{biometric sub-references} evolving with various
  template update methods, but authentication will be done with a
  unique biometric authentication method (whereas in
  multibiometrics, people use a multi-algorithm scheme when only one
  modality is used).
  Fig.~\ref{fig_schema} summarizes the proposed system (green
  area) for a
  system using two biometric sub-references per user (\emph{i.e.}, two
  different biometric template update systems evolve in parallel),
  and tab.~\ref{tab_diff_methods} presents the difference between self-update,
  co-update and hybrid-update.
  \begin{table*}[tb]
    \caption{Self-update, co-update and our hybrid-update behaves
    all differently}
    \label{tab_diff_methods}
    \centering
    \begin{tabular}{|p{.23\textwidth}|p{.23\textwidth}|p{.23\textwidth}|p{.23\textwidth}|}
      \hline
                   & \textbf{Self-update} & \textbf{Co-update} & \textbf{Hybrid-update} \\ \hline
		   \textbf{One modality}                 & yes              & no                                             & yes \\ \hline
		   \textbf{Several types of classifiers} & no               & yes (one per modality)                         & implementation choice (one per sub-reference or the same for all) \\ \hline
		   \textbf{Update decision source}       & classifier score & disagrement between the two classifiers scores & aggregated score \\ \hline
    \end{tabular}
  \end{table*}

  The defined system is independent of the other components of a
  template update system (pink and blue areas in
  fig.~\ref{fig_schema}).
  When a query is compared to the biometric meta-reference of the
  claimant, it is in fact compared to each biometric sub-reference.
  The scores are fused in order to obtain one aggregated score.
  We assume nothing on the update decision method;
  it can be based on a double thresholding method, a quality
  index, or anything else.
  With a double thresholding method, 
  the decision is taken on the aggregated score, so we do not know
  which of the biometric sub-reference is responsible
  of the update decision.
  This is not a problem because we are in a monomodal system
  and the comparison scores produced by the comparison to the
  various biometric sub-references must be highly correlated; we
  expect the opposite in co-update systems.
  When the biometric meta-reference must be updated, we update each of
  its biometric sub-reference (and not the less influent one as in
  co-update) using the accepted query and the
  template update method specific to each biometric
  sub-reference. 
  We expect this way to decrease the updating errors.
  Each couple of gallery update and method to compute the biometric reference can be replaced by an
  online classifier update~\cite{bhatt2011co}. 
  In this case, each online classifier must be different
  in order not to evolve identical biometric sub-references, as the aggregated
  score would be the same than the score of each online classifier.
%
  Different fusion rules and biometric sub-references updating
  methods can be used in this new update procedure.

  \subsection{Proposed Evaluation Metrics}

  There is a lack of evaluation metrics for template update 
  in the literature~\cite{giot2012performance}.
  Rattani \emph{et al.} use the ratio of impostors' samples
  present in the gallery after
  the update~\cite{rattani2008biometric}.
  It is used in an offline update
  procedure, whereas we want to evaluate the system in an online
  way (\emph{i.e.} the ratio of impostors can evolve after each
  query presentation to the biometric meta-reference).
  Poh \emph{et al.} explain how to estimate the authentication
  performance over time~\cite{poh2007method}.
  This procedure requires a dataset where samples are very well
  spread on a large time span, which is never the case when
  samples are acquired among various sessions (a lot of samples on
  a short time span and no samples at all on a long time span).

  To overcome these issues, we propose two evaluation metrics:
(i) the \emph{Impostor Update Selection Rate} (IUSR)
  which corresponds to the ratio of impostor's samples involved in the
  update process among all the tested impostor's samples;
(ii) the \emph{Genuine Update Miss Rate} (GUMR)
  which corresponds to the ratio of genuine's samples not involved in the
  update process among all the tested genuine's samples.
  Say, we have $N_t, N_i, N_g$ respectively the total number of tested
  samples, the number of tested impostor samples and the number of
  tested genuine's samples ($N_t=N_i+N_g$).
  Say, we have $U_i, U_g$ respectively the total number of
  impostor's
  samples selected in the updating process and the number of genuine's
  samples selected in the updating process.
  The error rates can be estimated as follows:
  \begin{equation}
    \widehat{IUSR} = \frac{U_i}{N_i}
  \end{equation}
  \begin{equation}
    \widehat{GUMR} = \frac{N_g-U_g}{N_g}
  \end{equation}

  In a system without template update mechanism, $IUSR=0$ and
  $GUMR=1$.
  The best template update systems tend to have a $IUSR$ as close
  as possible to 0 because the inclusion of impostor's samples is problematic
  as the biometric reference will attract more easily
  impostors.
  It also have the lower $GUMR$ as possible, but not equal to zero, because missing genuine
  samples can be a good thing when these samples are too noisy.
  Next section presents the configuration we have chosen to
  evaluate our new update procedure.

\begin{table}[!tb]
  \centering
  \caption{Experiment parameters.}
  \label{tab_parameters}
  \begin{tabular}{|p{2.7cm}|p{5cm}|}\hline
    \textbf{Parameter} & \textbf{Value} \\ \hline
    Modality              & Keystroke dynamics                      \\\hline
    Authentication method & Distance computing~\cite{demagalhaes2005pss}                      \\\hline
    Update decision       & Online double threshold semi-supervised \\\hline
    Update threshold      & Empirically fixed                                   \\\hline
    Update mechanism (of sub-references)     & None, sliding
    window, growing window \\\hline
    Number of sub-references & A biometric meta-reference is composed of
    2 biometric sub-references \\ \hline
    Fusion of references comparison distances & Mean value,
    minimum value (as we work with distance)\\
    \hline
    Aggregation combinations & (None, Sliding), (None, Growing),
    (Sliding, Growing)\\\hline
    Number of sessions    & 8 on DSL2009, 5 on GREYC2009            \\\hline
    Respect to chronology & Yes                                     \\\hline
    Presentation orders   & Random                                  \\\hline
    Input size& 30\% of impostors                                    \\\hline
    Evaluation computing  & Online (\emph{i.e.} joint
    adapt-and-test strategy per session~\cite{poh2009challenges})             \\\hline
    Evaluation metrics    & EER, FNMR, FMR, IUSR, GUMR
    (scores of current session, no error average with previous
    sessions)        \\\hline
  \end{tabular}
\end{table}

\section{Protocol}
  \label{sec_protocol}
  This section presents the precise configuration we have defined
  to evaluate our new template update procedure.
\subsection{Parametrization}
  Various parameters must be configured in order to evaluate a
  template update system and reproduce the study~\cite{seeger2011how,giot2012performance}.
Tab.~\ref{tab_parameters} summarises the various parameters used
in the experiment.
We have chosen to evaluate the proposed system on a behavioral
modality which presents more important temporal variations
than a morphological modality.
Among the available behavioral modalities, keystroke dynamics is
the one having the biggest datasets in term of number of
sessions.
The template update system is evaluated for each session using
only the scores computed during this session.
In order not to give over-optimistic results~\cite{giot2012performance}, we do not average
the performance of each session with the performance of the
previous sessions.
As the set of queries tested against a biometric reference
is randomly built, results can vary among the runs.
To cope with this variability, we launch the experiment 100 times and
present the averaged results.
  Two different keystroke dynamics datasets are used in order to
  validate the proposed work. 
  We use the DSL2009~\cite{killourhy2009cad} (51
  users, 400 samples per user, 8 sessions) and the
  GREYC2009~\cite{giot2009benchmark} (100 users, 60 samples per
  user, 5 sessions) databases.
  Due to the lack of space, only results on DSL2009 are
  presented, but conclusions are similar for GREYC2009.
  Session 1 is used for the enrolment stage, \emph{i.e.} to generate the
  initial biometric sub-references
(user's gallery size is the number of samples per session per user).
The other sessions serve to test and update the biometric
meta-reference.

\subsection{Configuration}
\label{sec_proposal_configuration}

  The template update methods are based on simple
  gallery update methods as in~\cite{kang2007crk}.
  Each time a gallery is modified, the associated biometric
  sub-reference is re-computed from scratch.
  Three gallery update methods are used: (i) \emph{none}, the gallery is not
  modified, there is no update; (ii) \emph{sliding window}, the selected
  query replaces the oldest
  sample of the gallery; (iii) \emph{growing window}, the selected query is added
  to the gallery.

  Three gallery aggregations methods are used: (a)
  \emph{parallel sliding},
  where one biometric reference is never updated, and the other one
  is updated with the sliding window;
  (b) \emph{parallel growing},
   where one biometric reference is never updated, and the other one
  is updated with the growing window;
  (c) \emph{double parallel}, where one biometric reference is updated
  using the sliding window, and the other one is updated
  using the growing window.
  Aggregation methods (a) and (b) produce two biometric
  sub-references following this rule: in the best case, one biometric
  sub-reference represents the behavior of the user at the initial enrolment, while the
  other represents its very last way of typing.

  Two score fusion methods are used: (1) the mean of the scores,
  and (2) the minimum value of the scores\footnote{The recognition
  method produces dissimilarity scores}, and
  each gallery aggregation method is used using each score
  fusion method.

  \subsection{Evaluation}
  The question is how to qualify if an updating system performs well ?
  We will see that, for keystroke dynamics
  systems, using no update
  results in a FNMR reduction with time (keystroke dynamics must be
  one of the rare biometrics having such behavior, because of the typing
  habituation, but we think that the FNMR is expected to increase
  after several additional sessions) and an increasing FMR (impostors are also
  expected to type better the password).
  A good update system is a system where FNMR and FMR both
  decrease (or remain stable) over time.
  In addition to these measures, we also present the IUSR and
  GUMR which provide information on the updating errors.
  The update decision is based on the similarity score, so
  FNMR/IUSR and FMR/FGMUR
  errors can be correlated.
  The EER is also used because of its ease of reading.

  \begin{figure}[!tb]
    \centering
    \includegraphics[width=\linewidth,ext=.pdf,type=pdf,read=.pdf]{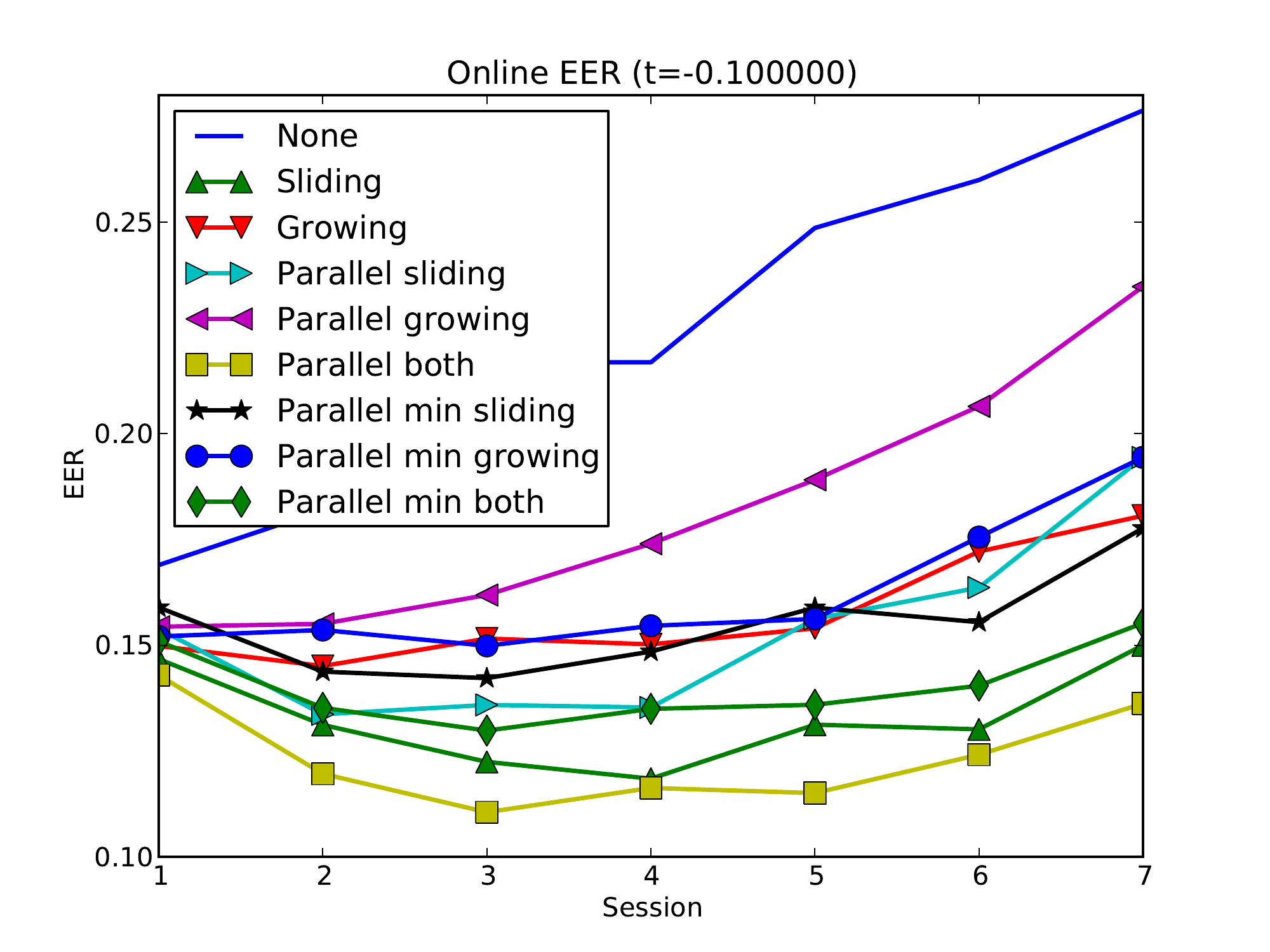}
    \caption{EER over sessions for each of the template update
    systems.}
    \label{fig_eer}
  \end{figure}

\section{Experimental Results}
\label{sec_results}

  \begin{figure*}[!tb]
    \centering
    \includegraphics[width=0.45\linewidth,ext=.pdf,type=pdf,read=.pdf]{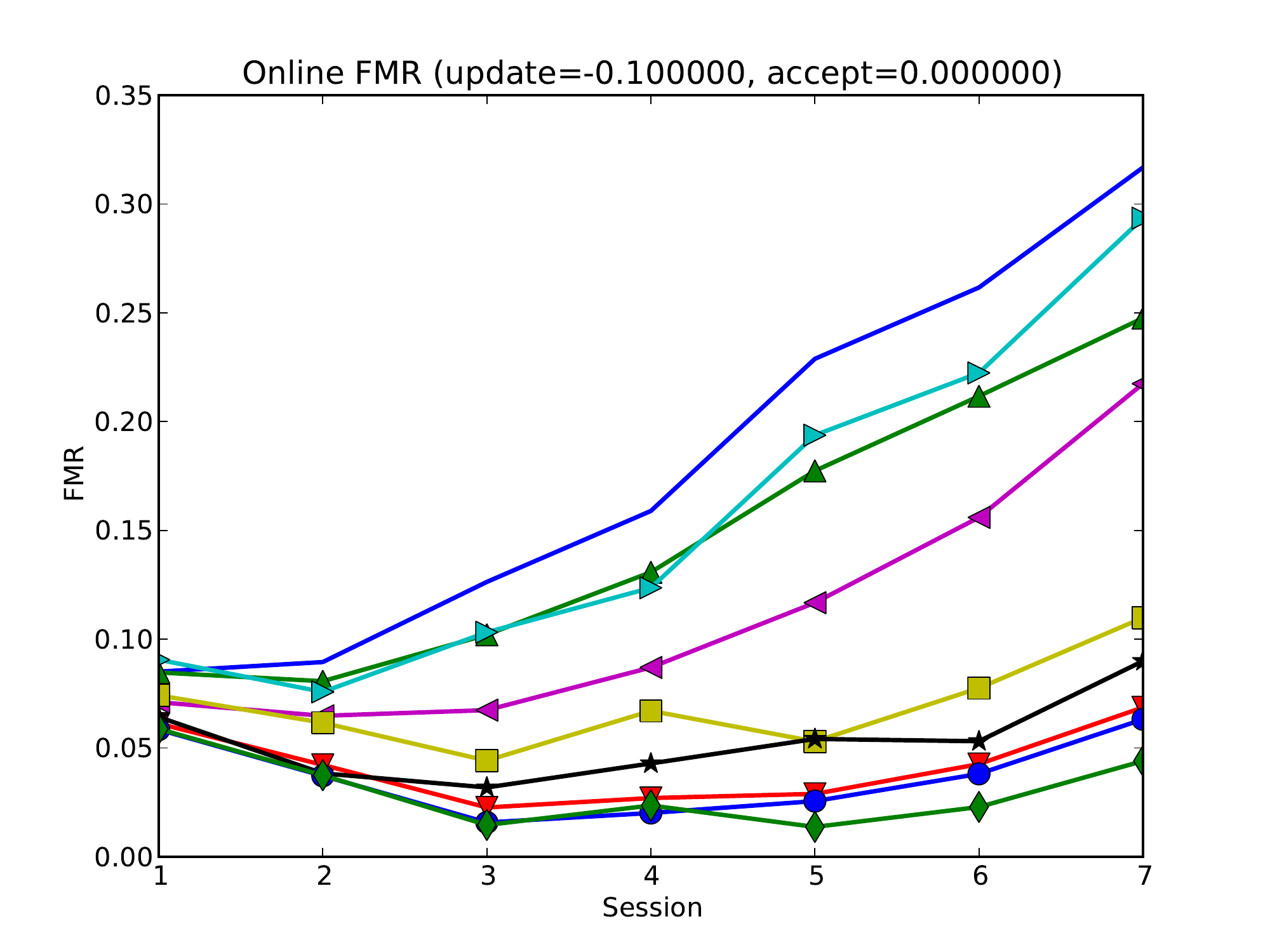}
    \includegraphics[width=0.45\linewidth,ext=.pdf,type=pdf,read=.pdf]{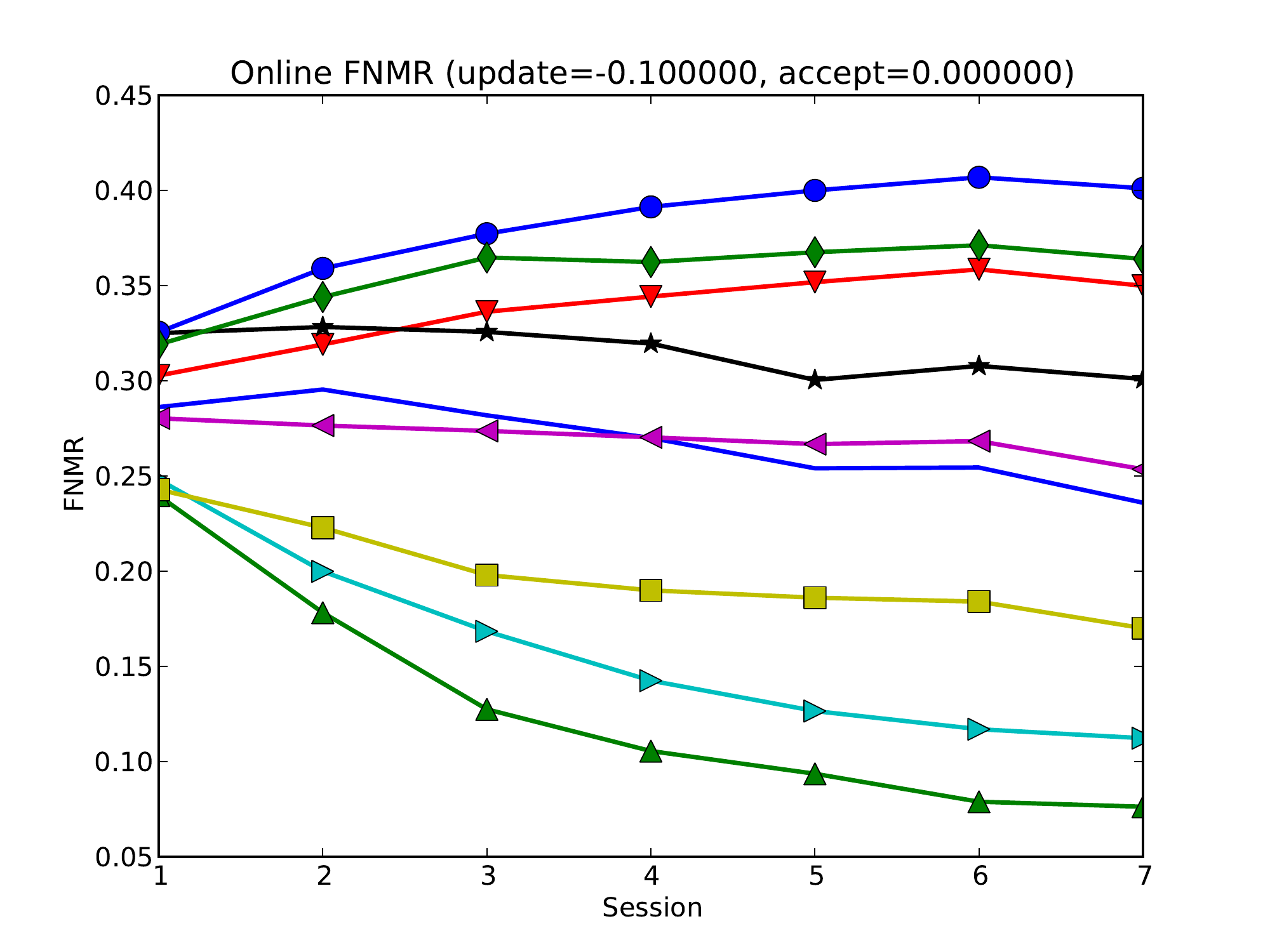}
    \caption{FNMR and FMR over sessions for each of the
    template update systems (accept threshold of 0.0, update threshold of -0.1).}
    \label{fig_fnmr_fmr}
  \end{figure*}

\begin{figure*}[!tb]
    \centering
    \includegraphics[width=0.45\linewidth,ext=.pdf,type=pdf,read=.pdf]{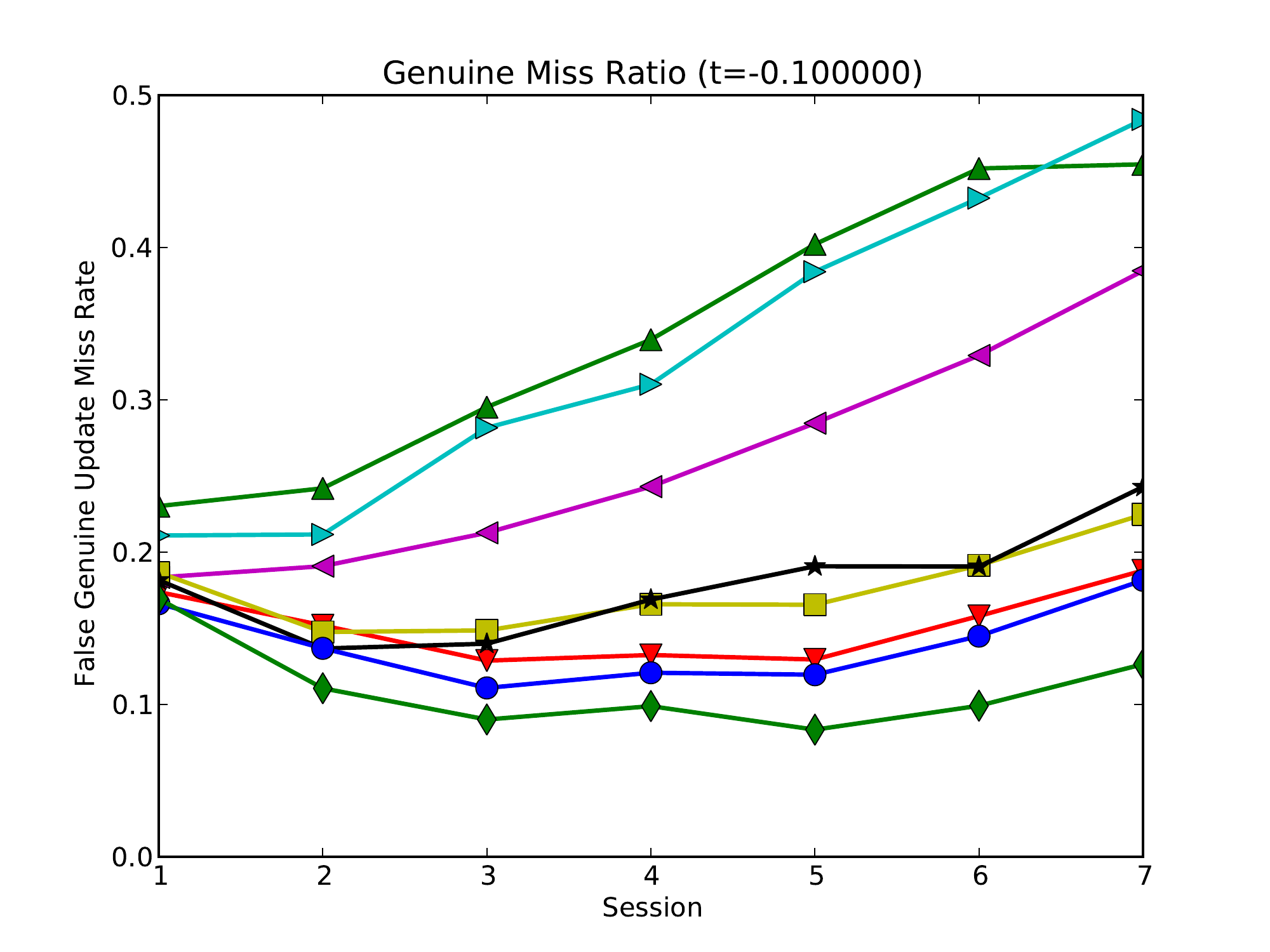}
    \includegraphics[width=0.45\linewidth,ext=.pdf,type=pdf,read=.pdf]{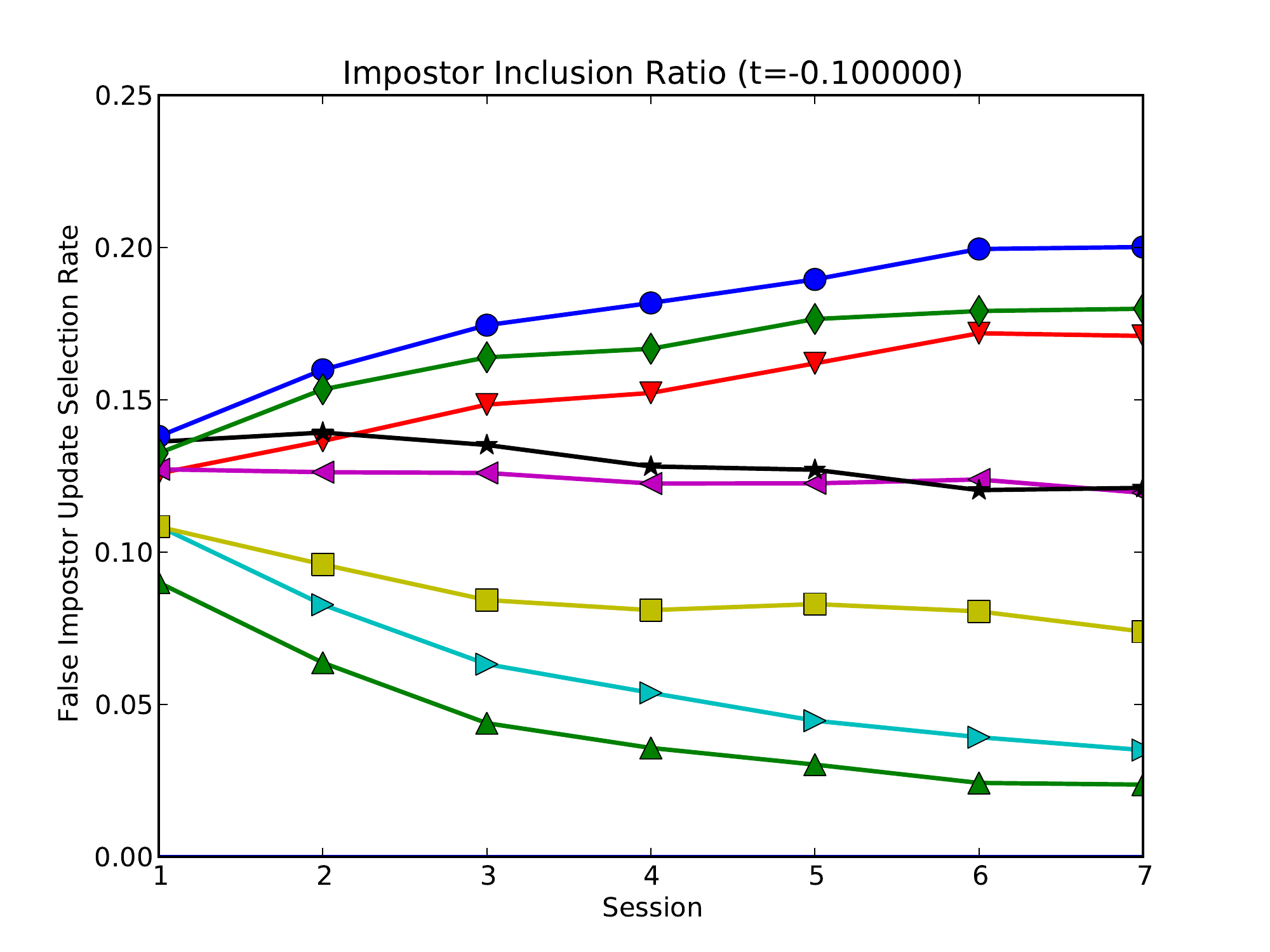}
    \caption{Update Error over sessions (update threshold of -0.1).}
    \label{fig_update_errors}
  \end{figure*}
 
  \begin{table*}[!tb]
    \centering
    \caption{Manual ranking of each method among various criteria.
    Top three methods using the EER in the ranks sum are in bold.
    Top three methods without EER are in italics.}
    \label{tab_ranks}
    \begin{tabular}{|l|lllll|ll|ll|} \hline
      \textbf{Method} &
      \textbf{FMR} &
      \textbf{FNMR} & 
      \textbf{EER} &
      \textbf{FISU} &
      \textbf{GMN} &
      \textbf{Score} &
      \textbf{Rank} &
      \textbf{Score without EER} & 
      \textbf{Rank}\\ \hline
\textbf{\emph{Parallel both}}*     & 5 & 3 & 1 & 4 & 4 & 17 & \textbf{1} & 16 & \textit{2} \\ \hline
\textbf{\emph{Parallel min both}}* & 1 & 8 & 3 & 8 & 1 & 18 & \textbf{2} & 15 & \textit{1} \\ \hline
\textbf{Sliding}                  & 7 & 1 & 2 & 2 & 8 & 23 & \textbf{3} & 21 & 4 \\ \hline
Parallel sliding*                  & 8 & 2 & 4 & 3 & 7 & 24 & 4          & 21 & 4 \\ \hline
\emph{Growing}                    & 3 & 7 & 6 & 7 & 3 & 26 & 5          & 19 & \textit{3} \\ \hline
Parallel min sliding*              & 4 & 6 & 5 & 6 & 5 & 26 & 5          & 21 & 4 \\ \hline
Parallel growing*                  & 6 & 4 & 8 & 5 & 6 & 29 & 7          & 21 & 4 \\ \hline
Parallel min growing*              & 2 & 9 & 7 & 9 & 2 & 29 & 7          & 22 & 8 \\ \hline
None                              & 9 & 5 & 9 & 1 & 9 & 33 & 9          & 24 & 9 \\ \hline
\end{tabular}
\end{table*}

The baseline scenario without template update is ``None'', and
the baseline scenarios with template update are the self-updates
with the ``Sliding'' and
``Growing'' gallery management ; they correspond to previous works
published in~\cite{giot2011analysis}.
Our new contributions in this paper are the other ones (
``Parallel sliding'',
``Parallel growing'',
``Parallel both'',
``Parallel min sliding'',
``Parallel min growing'',
``Parallel min both''
).

It is well known that decreasing the FNMR of a biometric system
correspond to increasing the FMR (and vice versa).
We can observe a similar behavior, linked to the time, on
fig.~\ref{fig_fnmr_fmr}.
Methods allowing decreasing the FNMR over time tend to increase the FMR
over time.
As the EER evolution cannot give us such kind of information (see
fig.~\ref{fig_eer}),
we think that, in opposition to previous
papers~\cite{rattani2011self,giot2012performance},
providing the EER of template update systems may be not be a good idea, and it
would be better to provide the FNMR and FMR in order to see their
difference of evolution.
In addition, when using a double threshold mechanism, the EER
threshold can be incompatible with the update threshold.

Looking on the fig.~\ref{fig_fnmr_fmr} (and not taking into
account fig.~\ref{fig_eer}, even if it could assert that too),
we see that the ``Parallel both'' method is the most appropriate.
It is not the best method in term of FMR nor FNMR, but it is
the sole method presents in the best methods each time.
As it seems to be a good compromise, we can say that using several
sub-references improves performances against using only one
(``growing'' and ``sliding'').
To assert this conclusion, we manually ranked each update method (sorted by global
  performance) on the following
  rates: FMR, FNMR, EER, FISUR, GUMR.
  For each update method, we sumed all the ranks of the various
  criteria and sort them.
  Tab.~\ref{tab_ranks} presents the ranking results.
  We have also computed the ranks without using the EER.
  We think we cannot trust the EER values, because (i) it may be
  hard to configure the system with the thresholds allowing to
  obtain the EER; (ii) the EER threshold may be incompatible with the
  one used for the update decision.
  Although ranks are different with the two ways of computing, the
  two bests and two worsts methods are the same.
  The two best methods are \emph{parallel both} and \emph{parallel
  min both}, which
  is the proposed method when we evolve in parallel two biometric
  sub-references using the growing window and the sliding window.
  It shows the benefit of the proposed method when evolving different
  biometric sub-references.
  The two worst methods are \emph{Parallel min growing} and
  \emph{None}.
  It is easy to understand.
  In the first case, there are two biometric sub-references: the initial
  reference which quickly becomes not representative and results in
  rejecting genuine samples and the growing window which can contains
  and keep a lot of impostor samples.
  This behavior is explained in fig.~\ref{fig_update_errors}
  where we see that this method is the one attracting the highest number
  of impostors.

  Fig.~\ref{fig_update_errors} shows that, for most methods,
  having a higher IUSR implies a lower GUMR (and vice-versa)
  except for \emph{parallel both}, which is never the best method,
  but is always in the top methods.
  It is the only method attracting not too many impostors and
  rejecting not too many genuine samples.
  The same behavior is observed on fig.~\ref{fig_fnmr_fmr}.
  Thus, being the method having not too much FMR and not much FNMR,
  the EER is low in comparison to other methods.

  Fig.~\ref{fig_update_errors} and fig.~\ref{fig_fnmr_fmr} show
  that there is a strong relationship between FNMR and GUMR, and FMR
  and IUSR. 
  This proves that to reduce the FNMR (respectively FMR), it is
  necessary to reduce the GUMR (respectively IUSR).
%
%
%
  There is one limit of the present evaluation procedure which is
  linked to the chosen update selection procedure.
  As we use a double threshold scheme, it is necessary to specify
  the two thresholds.
  Results would be different with thresholds performing badly.
  This may be an issue in an operational scenario (the optimum
  thresholds may be hard to obtain).
  A good practice would be to compute the threshold of a selected
  operational point using enrolment samples of all users, and
  compute the update threshold using it (using the EER threshold
  computed with first session 
  divided by 2 gives us similar results).

\section{Conclusion}
  \label{sec_conclusion}
  We have presented a hybrid template update method allowing to
  update several biometric references in parallel.
  The parallel evolution of biometric sub-references allows
  reducing the update error rates and the performance decreases over
  time in comparison to classic methods using one reference.
  The method has been validated on a template update system for
  keystroke dynamics on two datasets.
  One of the datasets contains 400 samples per users which is
  larger than most studies from the state of the art for template
  update of morphological modalities.
  We have shown that our scheme gives better performance than the
  classical ones (self-update with sliding or growing windows).
  Although the method has been evaluated in an online semi-supervised
  scenario, it could be used in offline scenarios or supervised
  scenarios too.
  The implementation uses two sub-references, but it would be
  useful to analyse if using more sub-references would improve
  the performances.
  It would be interesting to validate the proposition in other
  contexts and other modalities (signature for example), as well as
  with online classifiers instead of methods using a gallery and
  update decision methods.


%

\end{document}